\documentclass{article}
\usepackage[numbers]{natbib}
\usepackage {arxiv}

\usepackage{cite}
\usepackage{floatrow}
\usepackage{amsmath,amssymb,amsfonts}
\usepackage{algorithm}
\usepackage{algorithmic}
\usepackage{textcomp}
\usepackage{xcolor}
\usepackage{float}
\usepackage{booktabs}
\usepackage{multirow}
\usepackage{enumitem}
\usepackage{setspace}
\usepackage{graphicx}
\usepackage{array}
\usepackage{amsmath}
\usepackage{amssymb}
\usepackage{epsfig}
\usepackage{epstopdf}
\usepackage{subfigure}
\usepackage{comment}
\usepackage{tabularx}
\usepackage{threeparttable}
\usepackage{adjustbox}
\usepackage{soul}
\usepackage{dsfont}
\usepackage[hyphens]{url}
\usepackage{hyperref} 

\newcommand{\mds}[1]{\textcolor{black}{#1}}

\title{\LARGE \bf
Generalized Context in Cross Attention for Transfer Learning of Disjoint Tabular Data}

\author{Kazi F. Akhter,~
        Ibna Kowsar, \\
       Department of Computer Science\\
Tennessee State University\\
Nashville, TN, USA\\
\and 
{\bf Manar D. Samad}\\
Department of Computer Science\\
North Carolina Agricultural and Technical State University\\
Greensboro, NC, USA\\
\texttt{mdsamad@ncat.edu} \\
}

\begin{document}

\maketitle

\begin{abstract}

Unlike images and text, applying transfer learning to tabular data is challenging due to heterogeneity in feature types, structures, and semantics across disparate domains. Existing methods assume shared features across data tables to enable knowledge transfer between domains, which is unrealistic in practice. \mds{This paper introduces generalized context learning to remove the requirement of shared features across domains. The generalized context captured by transformer projection weights for $key$, $value$, and $query$ provides rule-based generalization rather than the domain-specific context conventionally learned from transformer activations. Projection weights for $key$ from the source domain interact with the weight for $query$ in the target domain to achieve Cross-domain Attention Transfer Learning (CATTLE) in a data-agnostic manner. Our experiments on ten pairs of disjoint source-target data sets show that CATTLE can learn generalized context from a single source data set and is rank-wise and statistically superior to nine state-of-the-art baselines, including machine learning, deep learning, and transfer learning methods using large-scale pre-trained models. CATTLE achieves the best average rank (2.9) and delivers a 3.7\% average AUROC gain over the baseline methods.} The CATTLE source code is available at~\footnote{\href{https://github.com/mdsamad001/CATTLE---cross-domain-transfer-learning-for-Tabular-Data}%
{https://github.com/mdsamad001/CATTLE---cross-domain-transfer-learning-for-Tabular-Data}}.
\end{abstract}


\section{Introduction}

Beyond what conventional machine learning can achieve, transfer learning has enabled a significant breakthrough in contemporary artificial intelligence.
Transfer learning leverages \mds {the general knowledge of the world} embedded in large, pre-trained models to adapt them in learning new domains with limited data. \mds {The conceptual framework of transfer learning} is established in computer vision and language models because images and text are conducive to transfer learning representations. Computer vision architectures, such as convolutional neural networks (CNNs) and vision transformers (ViTs), are successful in transfer learning because images provide generalizable, shared, and hierarchical feature representations that facilitate knowledge transfer across various domains~\citep{dosovitskiy2021an}. For example, large-vision models (e.g. ViT, VLM), trained with \mds {a diverse collection of} natural images, can be finetuned using monochromatic medical images for downstream disease classification tasks~\citep{Lee2025}. 

In contrast, data structured as \mds {rows and columns of} tables, also known as tabular data, \mds{have some fundamental distinctions from images and text. First, tabular data} contain a heterogeneous feature space~\citep{zhu2023xtab}, including numerical, categorical, ordinal and textual data types, unlike the homogeneous distribution of pixels and words in images and text. Consequently, traditional feature engineering and empirical selection of classifier models are still in practice for tabular data. Therefore, no single learning method consistently stands out as the best model for a diverse set of tabular data~\citep{rabbani2024attention}. \mds{Second, the tabular feature spaces from two distinct domains are often disparate and disjoint. For example, } banking transaction tables do not share plausible features with electronic health records data. Differences between diverse feature spaces pose a major challenge to acquiring broadly applicable knowledge, which is essential for effective transfer learning and for developing large foundation models. To this end, contemporary transfer learning approaches assume and exploit shared features between tabular data sets, typically within a single application domain, to ensure sufficient feature overlap~\citep{Wang2024, levin2023, onishi2023}. Transfer learning within the same domain assumes that the source and target data sets share similar feature distributions and contexts, enabling straightforward adaptation through model fine-tuning. In real-world scenarios, tabular datasets from different domains often vary greatly in their structure, feature types, and value distributions, and they rarely share overlapping features. \mds{A data-agnostic learning framework may alleviate the need for overlapping feature spaces and facilitate transfer learning between disjoint tabular data sets. This paper introduces a novel method that leverages transformer projection weights to learn a data-agnostic, general context, rather than relying on traditional attention-weighted data representations. While attention-weighted data representations provide a data-dependent means for cross-domain transfer learning, our data-agnostic approach facilitates cross-domain learning by engineering the projection weights pertaining to $key$, $value$, and $query$ representations. A single source tabular data set may be sufficient for data-agnostic learning without requiring a large number of data sets from numerous application domains. The subsequent sections provide additional motivation and background literature toward the contributions of this paper.}

\subsection{Motivation} \label{subsec: Motivation}
Tabular data are ubiquitous, thanks to real-world applications in countless real-world applications, including banking, business, healthcare, and government. Although deep learning currently dominates the field of artificial intelligence (AI), recent studies indicate that traditional machine learning (ML) methods continue to rival deep learning when applied to tabular data~\citep{Wang2021, Kohler2019, Smith2020, Shwartz-Ziv2022, Borisov2022_survey}, because \mds{the inductive bias necessary for a deep learning model to effectively learn heterogeneous tabular data remains an open problem.} In contrast, traditional ML does not offer several key advantages of deep learning approaches, such as representation learning, incremental learning, and transfer learning, which restricts our ability to construct large data-driven models that learn from heterogeneous tabular data sets. In this setting, transfer learning can support numerous domains where only a small amount of tabular data can be collected, for instance, rare diseases or costly laboratory procedures. Furthermore, transfer learning involves models pretrained using large labeled data sets. Although humans can identify and label images and text for constructing pretrained models, manually assigning labels to feature vectors in tabular data sets is neither straightforward nor strictly objective. Therefore, pretraining using a large volume of unlabeled tabular data for effective downstream finetuning and transfer learning is imperative, but challenging~\citep{levin2023}.

\subsection {\mds {Related work and our} contributions} \label{related_work}

Decision tree-based machine learning, such as XGBoost~\citep{chen2016xgboost}, \mds{has} dominated tabular data learning, despite the remarkable achievements of deep learning (DL) of image and text data~\citep{Grinsztajn2022}. Recent advances in deep tabular architectures, including TabNet~\citep{TabNet_Arik2021} and FT Transformer~\citep{FTT_Gorishniy2021}, outperform traditional machine learning models in some selective and large data sets with more than 10,000 samples. In practice, many tabular data sets have fewer than 1000 samples~\citep{rabbani-g-ceals-2024}, which may not fully leverage \mds{the strength of} deep representation learning. \mds{Recently, TabPFNv2~\citep{hollmann2025_tabpfnv2} has shown strong performance as a tabular foundation model on small data sets up to 10,000 samples. TabPFNv2 is trained exclusively on approximately 130 million synthetic data sets generated from a fixed prior defined by structural causal models. Inference is performed via in-context learning with a single forward pass, without dataset-specific fine-tuning or transfer learning. Another line of work uses text corpora from large knowledge bases like Wikidata for pretraining  CARTE~\citep{kim2024_carte} to represent tabular data with graph structure and TARTE~\citep{kim2025_tarte} using transformer models. TARTE outperforms CARTE on several benchmark tasks, underscoring the effectiveness of transformer models for tabular data. \mds{Alternatively, large language models (LLMs) have shown remarkable performance on zero-shot and few-shot learning of tabular data~\citep{han2024_featllm, hegselmann2023_tabllm}, which are largely constrained by token size limits and prone to subjectivity of prompt engineering. Simply finetuning LLMs on tabular data yields performance worse than traditional ML methods~\citep{rabbani2024_llm} due to the language-to-tabular representation barrier, which LATTLE~\citep{kowsar2025_llm} addresses by introducing a transformer adapter between language and tabular representations. We argue that a robust tabular-to-tabular learning paradigm is essential before breaking the barriers between language and tabular learning frameworks.}}

\mds {In contrast to the image and text domains, data augmentation and transfer learning approaches for tabular data sets of small-samples are still underdeveloped~\citep{onishi2023}.} Transfer learning studies have shown that fine-tuning a pretrained model yields better performance on tabular data than using models without pretraining~\citep{cheng2024why, Abrar2022Perturb}. However, pretraining and finetuning tasks in many studies are performed on subsets of the same tabular data set. Similarly, TTNet~\citep{Li2022ttnet} performs supervised pretraining followed by finetuning on another subset of the same \mds {tabular} data set. \mds{In practice, transfer learning involves pretrained models developed from disparate domains to learn general knowledge about the world. However, learning incrementally from tabular data of disparate domains remains challenging due to disjoint feature spaces.}

Several recent studies have investigated transfer learning for tabular data using a pair of source and target data sets. The pretraining step in similar studies learns \mds{shared representations of the source data to facilitate downstream learn of a target data set with limited samples.} TransTab~\citep{Wang2024} performs transfer learning between two different source and target data sets with several common features. An overlap in the feature space helps align the source and target \mds{data representations}, enabling effective knowledge transfer between tables. However, tabular data sets from disparate domains in the real world are unlikely to share features. Levin et al.~\citep{levin2023} present a transfer learning method inspired by the need to predict rare diseases using limited patient samples. A transformer-based framework is pretrained on a large source data set containing many patient samples for a multiclass disease classification task. The pre-trained model is fine-tuned using limited patient samples for a rare disease prediction task. The rare disease data set includes all the features used in the upstream multiclass disease classification in addition to those specific to rare diseases. The feature \mds{overlaps} between the source and target data within the same application domain may not represent the broader context necessary for cross-domain knowledge transfer.

Recent studies present transfer learning between distinct domains and table schemas. The Cross-table Masked Modeling (CM2) framework~\citep{Ye2024} is proposed to facilitate large-scale model pretraining using 2000 source data sets with varying schemas, feature types, and distributions. However, the 16 target data sets used for downstream classification share features and domains with their source counterparts. XTab~\citep{zhu2023xtab} is pre-trained by multiple source data sets using separate embedding layers for individual tables while simultaneously learning common feature representations using a shared transformer. Despite multiple source data sets for pre-training, XTab struggles to outperform traditional machine learning models (e.g., XGBoost~\citep{chen2016xgboost}) in downstream target data classification. Therefore, an effective \mds{strategy is necessary} to achieve robust transfer learning of tabular data across disparate domains. In computer vision, Seo et al.~\citep{Lee2024stochca} have proposed stochastic cross-attention (StochCA) learning using a pre-trained vision transformer (ViT). The StochCA method uses a pretrained ViT to obtain $key$ and $value$ representations for the target data. To obtain cross-attention between source and target, they perform a scaled dot product between the $key$ representation of the pretrained ViT and the $query$ representation of the downstream ViT. Although cross-attention works seamlessly for multi-source images due to homogeneous feature representations, it is not trivial for a heterogeneous feature space across disparate tabular data sets. \mds {Therefore, data-dependent (i.e., based on $key$ and $value$ representations) cross-attention mechanisms for vision and language models may not be the optimal choice for tabular data. Even to enable data-dependent learning for cross-domain knowledge transfer, existing work relies on the assumption that tabular data sets must share feature spaces. We argue that knowledge transfer across heterogeneous domains can be impeded by disjoint tabular data representations, potentially undermining the effectiveness of transfer learning. Transfer learning performance may be suboptimal because the target domain can overwrite or negate the knowledge originating from the source domain.}

To this end, we propose a novel generalized context necessary for learning cross-domain attention in tabular transfer learning (CATTLE). The generalized context is derived from the projection weights of a pretrained transformer, rather than relying on domain-specific $key$, $value$, and $query$ activations. To \mds {the best of our} knowledge, this is the first paper to introduce generalized context learning for tabular data sets, especially for transfer learning across domains without shared features. The key contributions of this work are as follows. First, this paper introduces a novel transfer learning \mds{frameworks for} disjoint tabular data sets of disparate domains without requiring shared feature spaces. Second, a novel method for learning cross-attention is introduced at the projection weight level of the transformer attention layer instead of using $key$, $value$, and $query$ activations. Third, the proposed generalized context can be obtained using a single tabular data set without requiring large-scale data sets from different domains. Fourth, the generalized context is data-agnostic, i.e., independent of the source-specific domain knowledge, but achieves state-of-the-art performance on downstream target data sets.

The remainder of the paper is organized as follows. Section~\ref{sec: method} presents the preliminaries on attention-based computing, \mds{conceptualize the generalized context,} and introduces the data-agnostic cross-attention framework for tabular transfer learning. Section~\ref{sec: experiments} discusses data sets, experiments, and evaluation methods.  Section~\ref{sec: results} compares the performance of our method with the state-of-the-art baselines and presents relevant ablation studies. Section~\ref{sec: Discussion} summarizes the main findings, provides insight into the results, and outlines limitations to suggest future work. The paper concludes in Section~\ref{sec: Conclusion}.

\section {Methodology} \label{sec: method}

\subsection{Preliminaries \mds {on attention}} \label{prelim}

A tabular data set $X \in \mathbb{R}^{N \times d}$ is structured in $N$ row samples and $d$ column features of varying types, where $d = a + b$ with $a$ categorical and $b$ numerical features. A transformer model treats each feature as a token and transforms the embedding ($E_i$ $\in \mathbb{R}^{d_m}$) of the individual token $i$ into a context vector ($Z_i$ $\in \mathbb{R}^{m}$), after learning between-feature attention as follows. The between-feature attention mechanism uses three trainable weights related to $query$, $key$, and $values$ ($W_q$, $W_k$, and $W_v$) of $\mathbb{R}^{m \times d_m}$  to project each token embedding ($E_i$) into its corresponding query ($q_i$), key ($k_i$), and value ($v_i$) vectors of $\mathbb{R}^{1 \times m}$, respectively, as shown in Equation \ref{eq-qkv}.
\begin{equation}
q_i = E_i W_q,\quad
k_i = E_i W_k,\quad
v_i = E_i W_v
\label{eq-qkv}
\end{equation}
The attention score ($\omega_{ji}$) for the feature $j$ to the feature $i$ is obtained using the scaled dot product of the query vector of $j$ ($q_j$) with the key vector of $i$ ($k_i$), where $\omega_{ji} \neq \omega_{ij}$. Attention scores are normalized using the dimension of the key vector ($d_k$) and the softmax function to obtain the corresponding attention weight $\alpha_{ji}$, as shown in Equation \ref{eq-weighted_sum}. \mds {In the context of tabular data, the attention weight $\alpha_{ji}$ quantifies how much the $query$ column $j$ should attend to the $key$ column $i$ for between-column contextual representation learning.} The context vector ($Z_j$) of the query feature $j$ is the attention-weighted sum of the $value$ vectors ($v_i$) corresponding to all features (columns), \mds{including feature $j$} (Equation \ref{eq-weighted_sum}). \mds{In other words, $Z_j\in \mathbb{R}^{m}$ is the representation of column $j$ in the context of all columns in the tabular data set.}
\begin{equation}
   \alpha_{ji} = softmax \left[ \frac{(q_j^Tk_i)}{\sqrt{d_k}}\right] , Z_j = \sum_{i=1}^{d} \alpha_{ji} v_i\label{eq-weighted_sum}
\end{equation}
\mds{Here, $j$ is the index of the $query$ feature and $i$ indexes the $key$ feature on which the $query$ feature attends. When $j$ = $i$, $\alpha_{jj}$ represents the attention of the feature $j$ to itself.} Transformers with $h$ attention heads result in a context vector of $\mathbb{R}^{m_h}$, where $m_h = m \times h$, \mds{by concatenating context vectors from all attention heads.}

\subsection{Generalized context via attention}

\mds{The representation of entity $j$ ($query$) in the context of entity $i$ ($key$) can be defined using the attention-weighted context vector ($Z_{ji}$) inspired by Equation~\ref{eq-weighted_sum}. However, the context is directly derived from data representations ($key$ and $value$), which is expected to facilitate data-dependent knowledge transfer between domains ($i$ and $j$) via cross-attention~\citep{mijoski2026weights}. Instead of a data-dependent, domain-specific context, a data-agnostic, generalized context can be used to facilitate learning across domains that do not share explicit context or feature spaces. Generalized context aims to provide knowledge about how to learn a new context rather than context about a specific domain or entity. Therefore, a knowledge transfer strategy independent of domain and data representations may serve to learn a generalized context for tabular data. Intuitively, the generalized context necessary to learn from data can be obtained from projection weight matrices related to $key$, $values$, and $query$ instead of their data representation vectors~\citep{chan2022transformers}. Projection weight matrices trained by a single source tabular data are assumed to have the necessary knowledge about how to project tabular data in general, without being specific to a particular domain.}

\subsection{Pretraining using source data} \label{lab:pretrain}

\begin{figure*}[t]
\centerline{\includegraphics[width=1.3\textwidth]{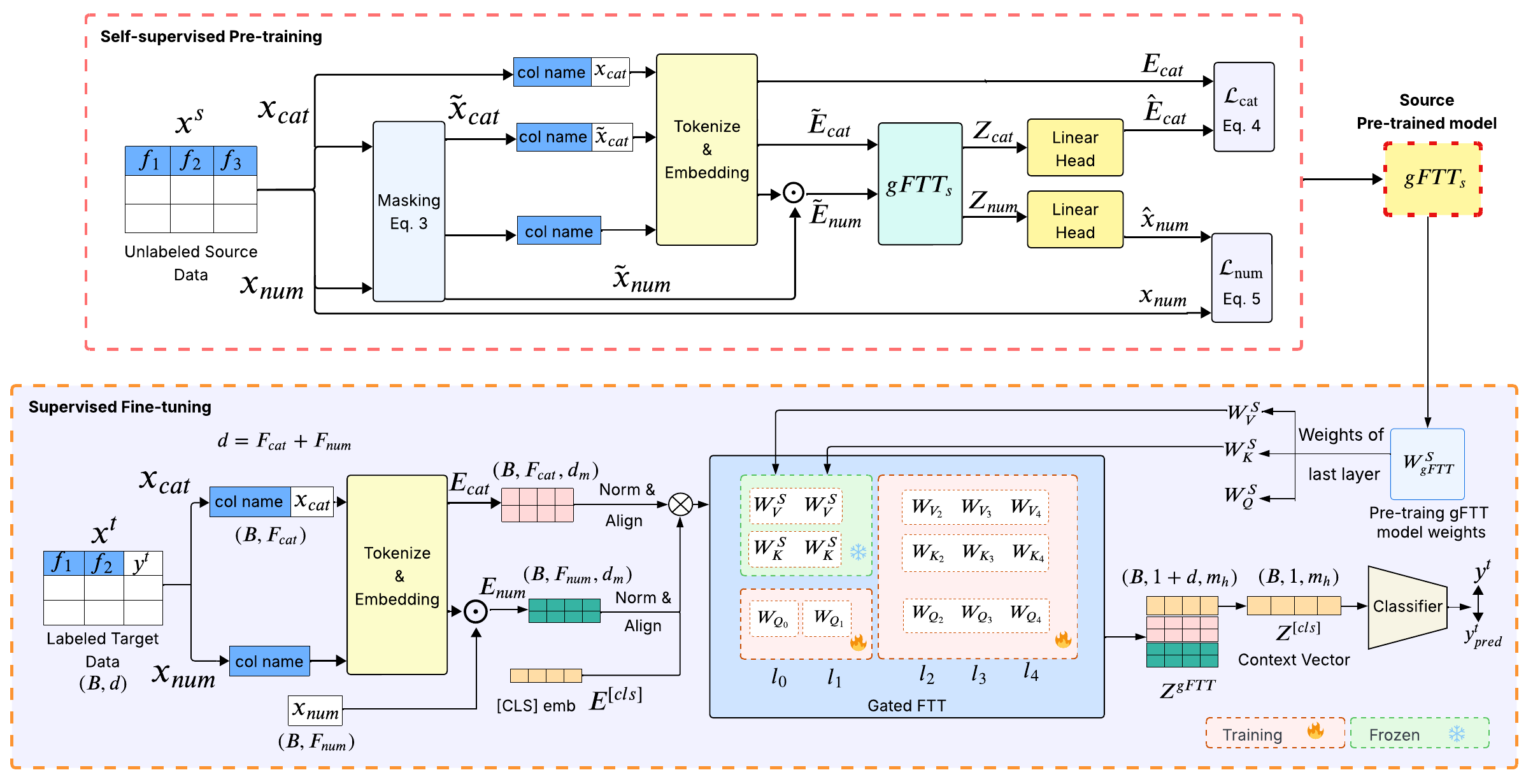}}
\vspace{-10pt}
\caption{Overview of the CATTLE framework. Attention-related weights from a pretrained Gated Feature Tokenizer Transformer (gFTT) trained on source data are transferred into a newly initialized gFTT, which is then finetuned using target data. Emb. = Embedding, Cat. = Categorical, and Num. = Numerical features.}
\label{fig-cattle}
\end{figure*}

Transfer learning involves pretraining on source data, followed by finetuning on target data to transfer knowledge. We obtain model pre-training separately by self-supervised and supervised learning for comparison. Self-supervised pre-training randomly masks a subset of data values and trains the model to reconstruct them from the unmasked values. A binary mask $M \in \{0,1\}^{N \times d}$ identifies whether or not each data element $M_{ij}$ is to be masked. The masking decision is made independently for each feature value using a Bernoulli distribution parameter $P_{\mathrm{mask}}$. Each mask entry $M_{ij}$ is set to 1 (masked) with a probability of $P_{\mathrm{mask}}$ and 0 (unmasked) with a probability of $1 - P_{\mathrm{mask}}$. \mds{ Ye et al.~\citep{Ye2024} have tested the masking ratio (Pmask) in a self-supervised cross-tabular learning framework like ours. They have empirically found that a masking ratio of 35\% yields the best performance among other values. Notably, when the mask ratio becomes too high, it discards an excessive number of feature values, which in turn harms self-supervised data reconstruction. A smaller masking ratio fails to introduce enough randomness into the data, making it difficult for self-supervised learning to capture meaningful relationships. } The value of a masked entry ($M_{ij} = 1$) is replaced with a \texttt{[MASK]} token, otherwise, the value is preserved, as shown in Equation~\ref{eq-1}.
\begin{equation}
\tilde{x} = x \odot (1 - M) + \texttt{[MASK]} \odot M
\label{eq-1}
\end{equation}
The operator $\odot$ denotes element-wise multiplication. The formulation in Equation~\ref{eq-1} retains the original feature value when $M_{ij} = 0$, and substitutes it with the masking token \mds{[MASK] when} $M_{ij} = 1$. The result is a partially observed input that preserves structure while forcing the model to reconstruct masked entries using the available context. To prevent complete information loss, we ensure that each sample has at least one unmasked feature. When all features in a row are masked, we explicitly unmask a fixed column in that row. 

The reconstruction of masked entries during pretraining is achieved differently for numerical and categorical features, as shown in Figure~\ref{fig-cattle}. The categorical feature name (“vehicle”) and the value (“sedan”) are tokenized to obtain the corresponding embeddings from a pretrained BERT model, which are then concatenated as a categorical feature embedding ($E^{cat}$). In contrast, the embedding of a numerical feature name is multiplied by the numerical value to obtain the embedding of the numerical feature ($E^{num}$). The embeddings for the numerical ($E_{num}$) and categorical ($E_{cat}$) features are concatenated to represent individual samples for pre-training the Gated Feature-Tokenized Transformer (gFTT)~\citep{wang2022transtab}. The reconstruction loss for categorical features is obtained using the cosine similarity between the reconstructed embedding vector ($\hat{E}^{cat}_{ij}$) and the input embedding $E^{cat}_{ij}$.  
\begin{equation}
\mathcal{L}_{\mathrm{cat}} = \frac{1}{n} \sum_{i=1}^{n} \sum_{j=1}^{a} m_{ij} \cdot \left( 1 - \mathrm{sim}(\hat{E}^{cat}_{ij}, E^{cat}_{ij}) \right)
\label{eq:cat_recon}
\end{equation}
Here, $m_{ij}$ represents an element of the binary mask matrix $M$, which is set to 1 to denote a masked value \mds{to be reconstructed}, otherwise zero for unmasked values. The loss in Equation~\ref{eq:cat_recon} maximizes the sum of similarity between the reconstructed and input embeddings of $a$ number of categorical features across $n$ samples.  For numerical features, the model reconstructs the masked values ($\hat{x}_{ij}$) from the context vectors $Z$. The mean-squared reconstruction error is computed across the $n$ samples and $b$ numerical features as follows.

\begin{equation}
\mathcal{L}_{\mathrm{num}} = \frac{1}{n} \sum_{i=1}^{n} \sum_{j=1}^{b} m_{ij} \cdot \| x_{ij} - \hat{x}_{ij} \|^2
\label{eq:num_recon}
\end{equation}
\mds{The sum of the loss values of all features grows with the number of features and should be normalized accordingly.} Therefore, the self-supervised feature reconstruction loss is obtained by \mds{separately normalizing the loss terms for categorical and numerical variables}, as shown in Equation~\ref{eq:total_loss}.
\begin{equation}
\mathcal{L}_{self} = \frac{1}{a} \mathcal{L}_{\mathrm{cat}} + \frac{1}{b} \mathcal{L}_{\mathrm{num}}
\label{eq:total_loss}
\end{equation}
\mds{It is important to note that categorical and numerical features undergo different encoding procedures, and consequently, contribute differently to model training. Therefore, we separately normalize the loss components for categorical and numerical features so that each type contributes in a balanced way. Using a common denominator like the total feature count (a+b) ignores how much each feature type contributes proportionally, which can cause one feature type to dominate the overall loss and diminish the influence of the other type. }
In supervised pretraining, the gFTT model is trained to predict the class labels of the source dataset using cross-entropy loss. We hypothesize that, for learning generalized context and obviating the need to annotate tabular data, self-supervised learning would be superior to supervised pretraining.

\subsection{\mds{Proposed} cross-attention for generalized context}

\mds{Transformer activations are commonly used in the literature to learn domain-specific context between the data representations of two comparable data sets~\citep{wang2022transtab, Ye2024, zhu2023xtab} (Equation~\ref{eq-weighted_sum}). However, disjoint tabular data from disparate domains lack a direct context that can be intuitively learned from activations or data representations. Instead of transformer activations, we argue that transformer projection weights can serve as a general data-transformation mechanism (e.g., how to project and learn from data) independent of domain-specific data and context. Our proposed idea is consistent with the work of Chan et al. \citep{chan2022transformers}, which suggests that the information stored in transformer weights is associated with rule-based generalization, whereas domain context is largely derived from corresponding data examples and representations \citep{mijoski2026weights}. Therefore, we hypothesize that rule-based generalization via transformer projection weights is more suitable for transfer learning across disjoint tabular data sets than transformer activations. Our novel mechanism for learning generalized context using transformer projection weights is discussed below.}

The $gFTT_s$ \mds{transformer model, discussed in Section \ref{lab:pretrain}}, trained by the source data set yields three transformer weights, $W^s_q$, $W^s_k$, and $W^s_v$, related to $query$, $key$, and $values$ representations, respectively. Here, $W^s_k$ and $W^s_v$ project the input to the $key$-$value$ (Equation~\ref{eq-qkv}) representation pair about the source data. However, the source data representation is assumed to be distributed across transformer layers, with the highest layer accumulating the most comprehensive knowledge. The layer with the most comprehensive knowledge is further assumed to have the most generalized context. Therefore, $W^s_k$ and $W^s_v$ of the highest layer of the pre-trained $gFTT_s$ are used as a foundation at the lowest layers of a new $gFTT_t$ (Figure \ref{fig-cattle}). The new $gFTT_t$ is then trained using target data on the foundation of generalized context transferred from the pretrained model. The cross-attention in a generalized context is achieved using the pretrained weights ($W^s_k$ and $W^s_v$) and projection weight ($W^t_q$) pertaining to the $query$ representation of the target data. While $key$, $value$, and $query$ representations are specific to data domains, corresponding projection weights provide generalized context for \emph{data-agnostic} cross-attention.  

The downstream $gFTT_t$ model keeps the pre-training weights ($W^s_k$ and $W^s_v$) frozen to retain the generalized context in the foundation while updating all other weights in the upper layers, propagating to the final inference for the target data set. It is worth noting that the hierarchical representation of images distributed across layers of convolutional neural networks is well known. However, a similar distribution of tabular data has not been well investigated for transformer layers. Therefore, this paper empirically investigates the distribution of knowledge across transformer layers for tabular data. Accordingly, the proposed \emph{data agonistic} cross-attention is achieved by setting the general context foundation at the lowest two layers of $gFTT_t$ by the frozen weight pairs ($W^s_k$ and $W^s_v$) and simultaneously updating the $query$ weight $W^t_q$ using target data. Subsequently, the scaled dot product (Equation \ref{eq-weighted_sum}) between the $query$ and the $key$ representations of the target data, where $key$ is projected by the pre-trained $W^s_k$, yields cross-domain attention for transfer learning. The proposed method is presented in Algorithm~\ref{alg:crossattention}.

\begin{algorithm}[h]
\caption{\bf{C}ross-\bf{AT}tention \bf{T}ransfer \bf{LE}arning (CATTLE)}
\label{alg:crossattention}
\begin{algorithmic}
\STATE\textbf{Input:} Source data set $(X_s, y_s)$, Target data set $(X_t, y_t)$ \\
\STATE\textbf{Model: $gFTT_s$ (source), $gFTT_t$ (target)} 
\STATE\textbf{Output:} Cross-attention $gFTT_{\text{CA}}$ for transfer learning
\STATE \textbf{$gFTT_s$ Pre-training: Source Data Set}
\FOR{epoch $= 1$ to $n\_epoch$}
\IF{Pre-training == Self-Supervised}
\STATE $gFTT_s \gets gFTT_s(X_s, \tilde{X_s})$, Using Eq.~\ref{eq:total_loss}
\ELSE
\STATE $gFTT_s \gets gFTT_s(X_s, y_s)$
  \ENDIF   
\ENDFOR
\STATE $gFTT_s$ with $L$ attention layers $\{l_1, l_2,...,L \}$
\STATE $\{W^{l_1}, \dots, W^L\} \gets gFTT_s$, Attention weights
\STATE $\{W_q^L, W_k^L, W_v^L\} \gets W^L$, Last layer weights\\
\STATE \textbf{$gFTT_t$ Transfer Learning: Target Data Set}
\STATE Weight Replacement for Cross-Attention:
\STATE \quad $gFTT_t$ [$W_k$, $W_v$]$^{\{\ell_0, \ell_1\}}$ $\leftarrow [W_k^L, W_v^L]$
\FOR{epoch $= 1$ to $n\_epoch$}
    \STATE $gFTT_{\text{CA}} \gets gFTT_t(X_t, y_t)$ with $\{W_k, W_v\}^{\{\ell_0, \ell_1\}}_{Frozen}$
\ENDFOR
\end{algorithmic}
\end{algorithm}
\section {Experiments} \label{sec: experiments}

\subsection{Tabular data in cross-domain pairing}

We use 14 tabular data sets from various application domains, such as health, finance, manufacturing, software testing, and industrial design, from the OpenML~\citep{OpenML2013} repository. The data sets have varying mixes of numerical and categorical features, sample sizes ranging from 540 to 70000, and feature dimensions ranging from 8 to 76. A summary of the tabular data sets is presented in Table~\ref{table-datasets}. We select nine source and five target data sets to form ten source-target pairs for transfer learning. Unlike existing work on tabular transfer learning~\citep{Ye2024, zhu2023xtab, wang2022transtab}, each pair of data sets is selected under a strict condition of disjoint feature space, ensuring that there are no shared features while differing in size and semantics.

\begin{table*}[t]
\centering
\caption{Summary of source and target data sets for transfer learning.}
\label{table-datasets}
\scalebox{0.55}{
\begin{tabular}{lllcccccc}
\toprule
Type & OpenML ID & Data set & Domain & Samples & Features & Numeric & Categorical & Classes \\
\midrule
\multirow{9}{*}{\shortstack{Source\\Data sets}} 
& 14    & mfeat-fourier (MF)           & Shape Measurements   & 2000  & 76 & 76 & 0  & 10 \\
& 31    & credit-g (CG)                & Banking            & 1000  & 20 & 7  & 13 & 2  \\
& 38    & sick (SK)                    & Thyroid Disease    & 3772  & 29 & 7  & 22 & 2  \\
& 980   & optdigits (DG)              & Optical Digits   & 5620  & 64 & 64 & 0  & 2  \\
& 1504  & steel-plates-fault (SP)      & Manufacturing      & 1941  & 33 & 33 & 0  & 2  \\
& 40664 & car-evaluation (CE)          & Car Pricing        & 1728  & 21 & 0  & 21 & 4  \\
& 40701 & churn (CH)                   & Telecommunication  & 5000  & 20 & 16 & 4  & 2  \\
& 45547 & cardiovascular-disease (CD) & Heart Disease      & 70000 & 11 & 5  & 6  & 2  \\
& 45562 & seismic-bumps (SB)           & Hazard Monitoring  & 2584  & 18 & 14 & 4  & 2  \\
\midrule
\multirow{5}{*}{\shortstack{Target\\Data sets}} 
& 23    & cmc (CM)                     & Demographics       & 1473  & 9  & 2  & 7  & 3  \\
& 37    & diabetes (DB)                & Metabolic Disease  & 768   & 8  & 8  & 0  & 2  \\
& 54    & vehicle (VH)                 & Automotive         & 846   & 18 & 18 & 0  & 4  \\
& 1068  & pc1 (PC1)                     & Software Testing   & 1109  & 21 & 21 & 0  & 2  \\
& 6332  & cylinder-bands (CB)          & Industrial Design  & 540   & 39 & 18 & 21 & 2  \\
\bottomrule
 \end{tabular}}
\end{table*}

\subsection {Baseline methods}

Baseline methods are selected from two groups: 1) direct classification of the target data sets without transfer learning and 2) state-of-the-art transfer learning methods proposed for tabular data. The first group without transfer learning includes XGBoost and Logistic Regression (LR) as representative machine learning models, ResNet and MLP as standard DL methods, TabNet as a transformer with attention-based feature selection during training~\citep{TabNet_Arik2021}, and FT-Transformer (FTT) using self-attention to learn tabular data representations~\citep{FTT_Gorishniy2021}. The state-of-the-art transfer learning methods include TransTab~\citep{wang2022transtab}, XTab~\citep{zhu2023xtab}, and CM2~\citep{Ye2024}. TransTab performs transfer learning between the source and target data sets with overlapping features by aligning shared embeddings~\citep{wang2022transtab}. XTab utilizes multi-table pretraining to enhance generalization across diverse tabular data sets with minimal feature overlap~\citep{zhu2023xtab}. CM2 learns generalized feature representations by aligning statistical distributions across diverse and heterogeneous tables~\citep{Ye2024}. 
\subsection{Model implementation and evaluation}
Transfer learning follows a two-step process: pre-training and fine-tuning. For CM2 and XTab, we use their pre-trained models trained on a large collection of tabular data sets. When pretrained models are unavailable, we use our source data sets for pre-training. In contrast, traditional machine learning and deep learning baselines are trained and evaluated using target data sets without transfer learning. The source and target data sets are randomly split into 70:10:20 for training, validation, and testing, respectively. Data with 70:10:20 splits are randomly sampled ten times using ten random seeds to facilitate statistical comparison of model performance. All models are validated using Optuna~\citep{optuna_akiba}. In each of the 100 Optuna trials, a set of hyperparameter values is randomly sampled and evaluated on the validation set. The best model following the validation is used to report the target classification performance using the test data fold. 

Our proposed transfer learning approach uses a gated feature tokenized transformer (gFTT) with five layers, where each layer consists of eight attention heads. The feedforward network has a hidden layer of size 2048, ReLU activation, and dropout. The gFTT model hyperparameters remain at the default setting during pre-training. For transfer learning, the gFTT trained on the target data is tuned using Optuna. Classification performance is reported using accuracy and the area under the receiver operating characteristic (AUROC) curve. A win matrix is presented after statistically comparing the AUROC scores of a pair of methods using the Wilcoxon signed rank test and a significance level of $\alpha <0.05$.

\section{Results} \label{sec: results}
All experiments are performed on Ubuntu 22.04, powered by an Intel(R) Xeon(R) W-2265 CPU (24 logical cores) running at 3.70GHz, 64GB of RAM, and a Quadro RTX A4000 GPU with 16GB of video memory. The performance of the proposed CATTLE method is compared with the baselines using 1) the average rank based on AUROC scores and 2) the Wilcoxon signed rank statistical tests presented in a Win matrix. 

\begin{figure}[t]
\centering
\subfigure[Self-supervised pretraining (Data set: car-evaluation)] { \includegraphics[trim=0.0 0 0 0cm, clip, width=0.45\textwidth] {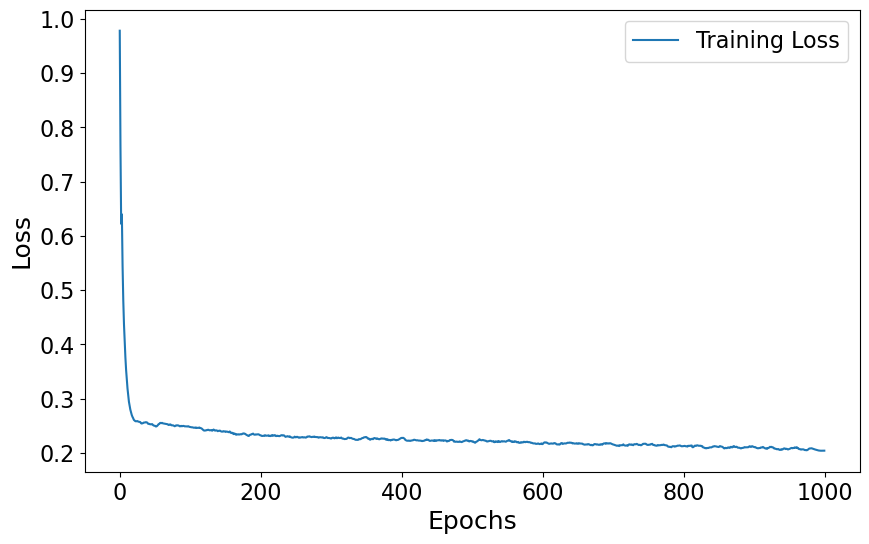}\label{fig:self-pretrain}}
\hspace{5pt}
\subfigure[Supervised pretraining (Data set: car-evaluation)] { \includegraphics[trim=0.0 0 0 0cm, clip, width=0.45\textwidth] {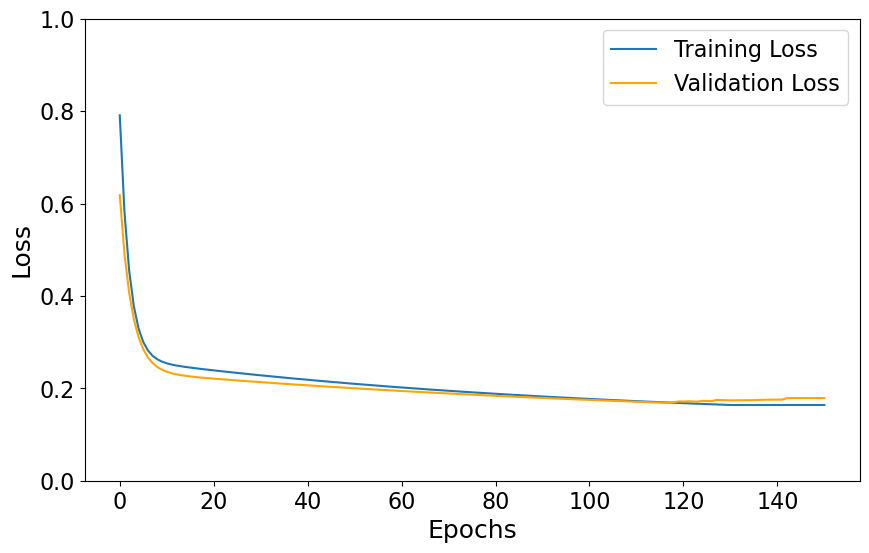}\label{fig:sup-pretrain}}
\subfigure[Finetuning after self-supervised pretraining (Data set: pc1)] { \includegraphics[trim=0.0 0 0 0cm, clip, width=0.45\textwidth] {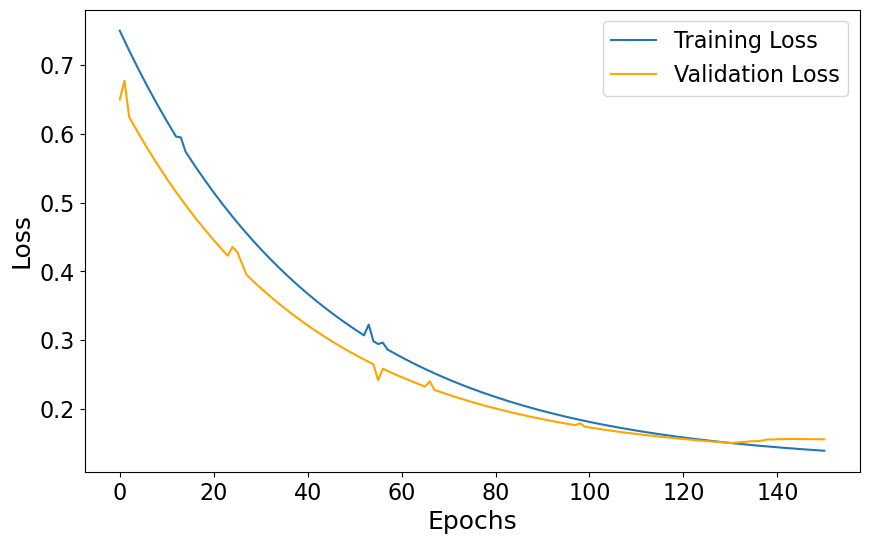}\label{fig:self-finetune}}
\hspace{5pt}
\subfigure[Finetuning after supervised pretraining (Data set: pc1)] { \includegraphics[trim=0.0 0 0 0cm, clip, width=0.45\textwidth] {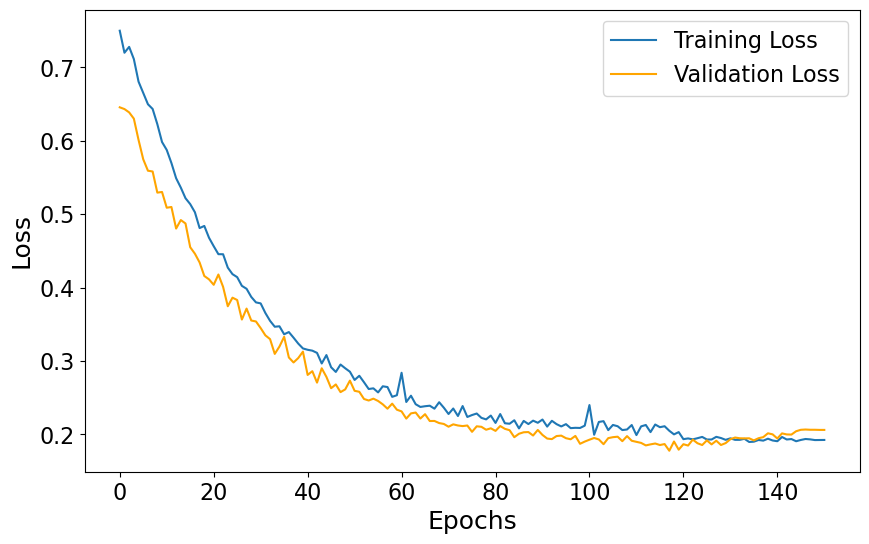}\label{fig:sup-finetune}}
\caption{Transfer learning loss curves: 
(a) self‐supervised pretraining, 
(b) supervised pretraining, 
(c) finetuning a self-supervised pretrained model using a target data set, 
and (d) finetuning a supervised pretrained model using a target data set.}
\label{fig:losses}
\end{figure}

\subsection{Model training and selection}

The upstream gFTT model with the default parameter settings is pre-trained on the source data. The self-supervised pre-training is performed for 1000 epochs with a batch size of 128 and a learning rate of 1e-4. The supervised pre-training uses 150 epochs, a batch size of 128, a learning rate of 3e-4, and a weight decay of 1e-2. \mds {It is noteworthy that self-supervised learning performs reconstruction of masked values in tabular data, which provides weaker training signals in backpropagation compared to label-supervised learning. The model is optimized to learn inter-feature dependencies for data reconstruction without explicit ground-truth targets. This observation is consistent with prior work showing that self-supervised methods require substantially more training iterations than their supervised counterparts due to slower convergence and higher computational demands~\citep{wang2021_selfsuperivsed}.  In contrast, explicit ground-truth labels, used as targets in supervised learning, yield faster convergence by providing a task-aligned gradient signal.} The loss curves for self-supervised and supervised pre-training are presented in Figure~\ref{fig:self-pretrain} and~\ref{fig:sup-pretrain}, respectively. \mds{The loss curves further confirm that supervised pretraining converges with 150 epochs, whereas self-supervised pretraining requires up to 1000 epochs of training.} In contrast, the downstream gFTT model is tuned using the hyperparameter space presented in Table~\ref{tab:hyperparameters}. Each Optuna trial randomly samples a set of hyperparameter values and evaluates them on validation data for 150 epochs. The convergence of training and validation losses for the target data set is demonstrated in Figure~\ref{fig:losses}, specifically for self-supervised pretraining (Figures~\ref{fig:self-finetune}) and supervised pretraining (Figure~\ref{fig:sup-finetune}).

\begin{table}[t]
\centering
\caption{Hyperparameter search space for all baselines and proposed methods.}
\label{tab:hyperparameters}
\scalebox{0.70}{
\begin{tabular}{ll}
\toprule
\textbf{Parameter} & \textbf{Distribution} \\
\midrule
\multicolumn{2}{l}{\textit{XGBoost~\citep{chen2016xgboost}}} \\ \midrule
\quad max\_depth          & UniformInt[1, 10] \\
\quad learning\_rate      & LogUniform[$\exp(-7)$, 1] \\
\quad n\_estimators       & UniformInt[100, 4000] \\
\quad subsample          & UniformFloat[0.2, 1.0] \\
\quad colsample\_bytree   & UniformFloat[0.2, 1.0] \\
\quad min\_child\_weight & LogUniform[$\exp(-16)$, $\exp(5)$] \\
\quad gamma              & LogUniform[$\exp(-16)$, $\exp(2)$] \\
\quad reg\_alpha         & LogUniform[$\exp(-16)$, $\exp(2)$] \\
\quad reg\_lambda        & LogUniform[$\exp(-16)$, $\exp(2)$] \\
\midrule
\multicolumn{2}{l}{\textit{Logistic Regression~\citep{kleinbaum2002logistic}}} \\ \midrule
\quad C & LogUniform[1e$^{-4}$, 1e$^{2}$] \\
\quad penalty & Categorical\{"l1", "l2"\} \\
\quad max\_iter & UniformInt[100, 1000] \\
\midrule
\multicolumn{2}{l}{\textit{MLP, ResNet, FT-Transformer~\citep{FTT_Gorishniy2021}}} \\ \midrule
\quad learning\_rate & LogUniform[1e$^{-5}$, 3e$^{-4}$] \\
\quad batch\_size & Int[32, 128] (step=32) \\
\quad weight\_decay & LogUniform[1e$^{-6}$, 1e$^{-2}$] \\
\quad hidden\_dropout\_prob & Categorical\{0.0, 0.1, 0.2, 0.3, 0.4\} \\
\midrule
\multicolumn{2}{l}{\textit{TabNet~\citep{TabNet_Arik2021}}} \\ \midrule
\quad mask\_type & Categorical\{"entmax", "sparsemax"\} \\
\quad learning\_rate & LogUniform[1e$^{-3}$, 1e$^{-1}$] \\
\quad cat\_emb\_dim & Int[8, 32] (step=8) \\
\quad gamma & Uniform[1.0, 3.0] \\
\quad batch\_size & Categorical\{64, 128, 256\} \\
\midrule
\multicolumn{2}{l}{\textit{XTab~\citep{zhu2023xtab}, TransTab~\citep{wang2022transtab}, CM2~\citep{Ye2024}}} \\ \midrule
\quad learning\_rate & LogUniform[1e$^{-5}$, 3e$^{-4}$] \\
\quad batch\_size & Int[32, 128] (step=32) \\
\quad weight\_decay & LogUniform[1e$^{-6}$, 1e$^{-2}$] \\
\quad hidden\_dropout\_prob & Categorical\{0.0, 0.1, 0.2, 0.3, 0.4\} \\
\quad warmup\_ratio & Categorical\{0.01, 0.05, 0.1\} \\
\midrule
\multicolumn{2}{l}{\textit{CATTLE}} \\ \midrule
\quad learning\_rate & LogUniform[1e$^{-5}$, 3e$^{-4}$] \\
\quad batch\_size & Int[32, 128] (step=32) \\
\quad weight\_decay & LogUniform[1e$^{-6}$, 1e$^{-2}$] \\
\quad hidden\_dropout\_prob & Categorical\{0.0, 0.1, 0.2, 0.3, 0.4\} \\
\quad warmup\_ratio & Categorical\{0.01, 0.05, 0.1\} \\
\bottomrule
\end{tabular}
}
\end{table}

\begin{table*} [t]
\centering
\vspace{-10pt}
\caption{Average area under the ROC curve (AUROC) and corresponding performance rank. \mds{Two-letter acronyms of the source and target data sets are taken from Table~\ref{table-datasets}. Traditional deep and machine learning methods for tabular data are trained and tested directly on target data sets without pretraining on the source.}}
\label{tab:cattle-result}
\scalebox{0.475}{
\begin{tabular}{ccccccccccccccc}
\toprule
& \begin{tabular}[c]{@{}c@{}}Logistic\\ Regression\end{tabular} & XGBoost & MLP & ResNet & \begin{tabular}[c]{@{}c@{}}FT-\\ Transformer\end{tabular} & TabNet & \multicolumn{2}{c}{XTab} & \multicolumn{2}{c}{CM2} & \multicolumn{2}{c}{TransTab} & \begin{tabular}[c]{@{}c@{}}CATTLE\\ Supervised\end{tabular} & \begin{tabular}[c]{@{}c@{}}CATTLE\\ Self-Supervised\end{tabular} \\
\midrule
Target &  &  &  &  &  &  & Source & & Source &  & Source &  &  &  \\
\midrule
\multirow{2}{*}{DB} & \multirow{2}{*}{0.822 (0.04)} & \multirow{2}{*}{0.810 (0.04)} & \multirow{2}{*}{0.798 (0.03)} & \multirow{2}{*}{0.818 (0.03)} & \multirow{2}{*}{0.788 (0.04)} & \multirow{2}{*}{0.643 (0.03)} & \multirow{10}{*}{\begin{tabular}[c]{@{}c@{}}52\\ AutoML\\ Data\\ sets\end{tabular}} & \multirow{2}{*}{0.834 (0.02)} & \multirow{10}{*}{\begin{tabular}[c]{@{}c@{}}Open\\Tabs\\ 2000 \\ Data\\ sets\end{tabular}} & \multirow{2}{*}{0.803 (0.03)} & CD & 0.799 (0.04) & 0.803 (0.04) & 0.807 (0.04) \\
& & & & & & & & & & & CG & 0.818 (0.04) & 0.808 (0.04) & 0.819 (0.04) \\
\multirow{2}{*}{VH} & \multirow{2}{*}{0.935 (0.01)} & \multirow{2}{*}{0.925 (0.01)} & \multirow{2}{*}{0.917 (0.01)} & \multirow{2}{*}{0.855 (0.01)} & \multirow{2}{*}{0.914 (0.01)} & \multirow{2}{*}{0.794 (0.10)} & & \multirow{2}{*}{0.935 (0.01)} & & \multirow{2}{*}{0.893 (0.01)} & MF & 0.928 (0.01) & 0.940 (0.01) & 0.933 (0.01) \\
& & & & & & & & & & & DG & 0.935 (0.01) & 0.932 (0.01) & 0.942 (0.01) \\
\multirow{2}{*}{CM} & \multirow{2}{*}{0.703 (0.02)} & \multirow{2}{*}{0.727 (0.02)} & \multirow{2}{*}{0.704 (0.02)} & \multirow{2}{*}{0.695 (0.03)} & \multirow{2}{*}{0.724 (0.03)} & \multirow{2}{*}{0.681 (0.08)} & & \multirow{2}{*}{0.721 (0.02)} & & \multirow{2}{*}{0.729 (0.02)} & CH & 0.726 (0.01) & 0.721 (0.03) & 0.729 (0.02) \\
& & & & & & & & & & & SK & 0.733 (0.02) & 0.726 (0.03) & 0.738 (0.02) \\
\multirow{2}{*}{PC1} & \multirow{2}{*}{0.826 (0.04)} & \multirow{2}{*}{0.834 (0.04)} & \multirow{2}{*}{0.825 (0.06)} & \multirow{2}{*}{0.681 (0.04)} & \multirow{2}{*}{0.810 (0.04)} & \multirow{2}{*}{0.843 (0.08)} & & \multirow{2}{*}{0.817 (0.05)} & & \multirow{2}{*}{0.845 (0.10)} & CE & 0.825 (0.04) & 0.842 (0.05) & 0.840 (0.08) \\
& & & & & & & & & & & SP & 0.812 (0.04) & 0.845 (0.04) & 0.813 (0.05) \\
\multirow{2}{*}{CB} & \multirow{2}{*}{0.816 (0.03)} & \multirow{2}{*}{0.847 (0.04)} & \multirow{2}{*}{0.603 (0.08)} & \multirow{2}{*}{0.594 (0.07)} & \multirow{2}{*}{0.677 (0.01)} & \multirow{2}{*}{0.687 (0.05)} & & \multirow{2}{*}{0.822 (0.03)} & & \multirow{2}{*}{0.824 (0.05)} & SP & 0.827 (0.03) & 0.829 (0.04) & 0.862 (0.04) \\
& & & & & & & & & & & SB & 0.825 (0.04) & 0.832 (0.04) & 0.850 (0.04) \\
\midrule
\begin{tabular}[c]{@{}c@{}}Avg.\\ Rank\end{tabular} & 5.0 (2.9) & 4.7 (1.6) & 8.1 (1.4) & 9.1 (3.0) & 8.5 (1.8) & 8.7 (3.5) & \multicolumn{2}{c}{4.6 (2.9)} & \multicolumn{2}{c}{4.7 (3.2)} & & 4.8 (2.4) & 3.9 (2.2) & 2.9 (2.4) \\
\midrule
\begin{tabular}[c]{@{}c@{}}Overall\\ Rank\end{tabular} & \multicolumn{1}{c}{7} & \multicolumn{1}{c}{4} & \multicolumn{1}{c}{8} & \multicolumn{1}{c}{11} & \multicolumn{1}{c}{9} & \multicolumn{1}{c}{10} & \multicolumn{2}{c}{3} & \multicolumn{2}{c}{5} & & \multicolumn{1}{c}{6} & \multicolumn{1}{c}{2} & \multicolumn{1}{c}{1} \\
\bottomrule

\end{tabular}}
\end{table*}

\begin{table*}[t]
\centering
\caption{Comparison of CATTLE with other baseline methods based on average classification accuracy. \mds{Two-letter acronyms of the source and target data sets are taken from Table~\ref{table-datasets}. Traditional deep and machine learning methods for tabular data are trained and tested directly on target data sets without pretraining on the source.}
}
\label{tab:cattle-result-acc}
\scalebox{0.475}{
\begin{tabular}{ccccccccccccccc}
\toprule
& \begin{tabular}[c]{@{}c@{}}Logistic\\ Regression\end{tabular} & XGBoost & MLP & ResNet & \begin{tabular}[c]{@{}c@{}}FT-\\ Transformer\end{tabular} & TabNet & \multicolumn{2}{c}{XTab} & \multicolumn{2}{c}{CM2} & \multicolumn{2}{c}{TransTab} & \begin{tabular}[c]{@{}c@{}}CATTLE\\ Supervised\end{tabular} & \begin{tabular}[c]{@{}c@{}}CATTLE\\ Self-Supervised\end{tabular} \\
\midrule
Target &  &  &  &  &  &  & Source & & Source &  & Source &  &  &  \\
\midrule
\multirow{2}{*}{DB} & \multirow{2}{*}{0.747 (0.04)} & \multirow{2}{*}{0.729 (0.04)} & \multirow{2}{*}{0.753 (0.03)} & \multirow{2}{*}{0.772 (0.03)} & \multirow{2}{*}{0.754 (0.04)} & \multirow{2}{*}{0.625 (0.05)} & \multirow{10}{*}{\begin{tabular}[c]{@{}c@{}}52\\ AutoML\\ Data\\ sets\end{tabular}} & \multirow{2}{*}{0.764 (0.02)} & \multirow{10}{*}{\begin{tabular}[c]{@{}c@{}}Open\\ Tabs\\ 2000\\ Data\\ sets\end{tabular}} & \multirow{2}{*}{0.734 (0.02)} & CD & 0.734 (0.04) & 0.735 (0.02) & 0.740 (0.04) \\
& & & & & & & & & & & CG & 0.735 (0.04) & 0.738 (0.05) & 0.746 (0.05) \\
\multirow{2}{*}{VH} & \multirow{2}{*}{0.736 (0.03)} & \multirow{2}{*}{0.753 (0.03)} & \multirow{2}{*}{0.754 (0.03)} & \multirow{2}{*}{0.721 (0.04)} & \multirow{2}{*}{0.767 (0.06)} & \multirow{2}{*}{0.574 (0.01)} & & \multirow{2}{*}{0.753 (0.02)} & & \multirow{2}{*}{0.664 (0.02)} & MF & 0.776 (0.03) & 0.774 (0.02) & 0.786 (0.02) \\
& & & & & & & & & & & DG & 0.766 (0.03) & 0.758 (0.02) & 0.773 (0.02) \\
\multirow{2}{*}{CM} & \multirow{2}{*}{0.512 (0.04)} & \multirow{2}{*}{0.551 (0.03)} & \multirow{2}{*}{0.526 (0.02)} & \multirow{2}{*}{0.536 (0.04)} & \multirow{2}{*}{0.583 (0.06)} & \multirow{2}{*}{0.766 (0.01)} & & \multirow{2}{*}{0.653 (0.03)} & & \multirow{2}{*}{0.552 (0.03)} & CH & 0.510 (0.02) & 0.534 (0.04) & 0.640 (0.03) \\
& & & & & & & & & & & SK & 0.558 (0.03) & 0.542 (0.04) & 0.647 (0.03) \\
\multirow{2}{*}{PC1} & \multirow{2}{*}{0.829 (0.04)} & \multirow{2}{*}{0.889 (0.01)} & \multirow{2}{*}{0.822 (0.01)} & \multirow{2}{*}{0.630 (0.01)} & \multirow{2}{*}{0.710 (0.04)} & \multirow{2}{*}{0.743 (0.04)} & & \multirow{2}{*}{0.717 (0.05)} & & \multirow{2}{*}{0.765 (0.10)} & CE & 0.918 (0.04) & 0.931 (0.05) & 0.928 (0.01) \\
& & & & & & & & & & & SP & 0.925 (0.01) & 0.932 (0.06) & 0.929 (0.01) \\
\multirow{2}{*}{CB} & \multirow{2}{*}{0.746 (0.03)} & \multirow{2}{*}{0.806 (0.04)} & \multirow{2}{*}{0.673 (0.06)} & \multirow{2}{*}{0.700 (0.05)} & \multirow{2}{*}{0.557 (0.02)} & \multirow{2}{*}{0.577 (0.04)} & & \multirow{2}{*}{0.722 (0.03)} & & \multirow{2}{*}{0.727 (0.05)} & SP & 0.778 (0.04) & 0.758 (0.04) & 0.789 (0.04) \\
& & & & & & & & & & & SB & 0.764 (0.04) & 0.743 (0.07) & 0.778 (0.04) \\
\midrule
\begin{tabular}[c]{@{}c@{}}Average\\ Rank\end{tabular} & 6.8 (2.5) & 5.0 (2.9) & 7.0 (2.2) & 8.1 (3.0) & 6.6 (3.8) & 7.9 (4.1) & \multicolumn{2}{c}{5.6 (2.9)} & \multicolumn{2}{c}{7.3 (1.9)} & \multicolumn{2}{c}{4.6 (3.0)} & 4.6 (2.7) & 2.4 (1.5) \\
\midrule
\begin{tabular}[c]{@{}c@{}}Overall\\ Rank\end{tabular} & \multicolumn{1}{c}{7} & \multicolumn{1}{c}{4} & \multicolumn{1}{c}{8} & \multicolumn{1}{c}{11} & \multicolumn{1}{c}{6} & \multicolumn{1}{c}{10} & \multicolumn{2}{c}{5} & \multicolumn{2}{c}{9} & & \multicolumn{1}{c}{3} & \multicolumn{1}{c}{2} & \multicolumn{1}{c}{1} \\
\bottomrule
\end{tabular}}
\end{table*}

\subsection {\mds{Comparison of model performance}} \label{ranking}

\mds{We report AUROC and accuracy (ACC) to evaluate and contrast model performance from two complementary viewpoints. AUROC reflects how well a model balances sensitivity and specificity over a range of classification thresholds, while ACC evaluates the proportions of true positives and true negatives at a single, fixed threshold. Therefore, the ranking of performance varies with the choice of performance metrics.}

Tables~\ref{tab:cattle-result} and Table~\ref{tab:cattle-result-acc} present the source-target data set pairs, average AUROC, and ACC scores after testing the target data sets. Notably, there is no single learning algorithm in practice that performs the best across all tabular data sets due to heterogeneity in data types and structures. Therefore, learning algorithms are ranked across data sets based on AUROC and ACC scores. Traditional machine and state-of-the-art DL methods for tabular data are not designed for transfer learning and, therefore, are trained and tested using target data. 

Traditional machine learning methods, particularly XGBoost and logistic regression, consistently achieve strong performance across most data sets. For example, the performance rank of XGBoost (AUROC: 4.7 (1.6), ACC: 5.0 (2.9)) is better than the best rank (AUROC: 8.1 (1.4), ACC: 6.6 (3.8)) among state-of-the-art DL methods, including ResNet, MLP, FT-Transformer~\citep{FTT_Gorishniy2021} and TabNet~\citep{TabNet_Arik2021}. This observation supports a similar finding that traditional ML methods remain competitive with state-of-the-art DL methods~\citep{Grinsztajn2022, rabbani2024attention} for tabular data. \mds{However, recent transfer learning methods for tabular data (XTab~\citep{zhu2023xtab}, CM2~\citep{Ye2024}, and TransTab~\citep{wang2022transtab}) consistently rank better than all of their DL counterparts without transfer learning based on AUROC scores. However, the average rank of CM2 based on ACC (AUROC: 4.7 (3.2), ACC: 7.3 (1.9)) is worst than two DL methods: MLP (AUROC: 8.1 (1.4), ACC: 7.0 (2.2)) and FT-Transformer (AUROC: 8.5 (1.8), ACC: 6.6 (3.8)). The TransTab method consistently ranks the best among transfer learning baselines for tabular data based on ACC scores (AUROC: 4.8 (2.4), ACC: 4.6 (3.0)), but falls short on AUROC scores compared to XTab ((AUROC: 4.6 (2.9), ACC: 5.6 (2.9)) and CM2 ((AUROC: 4.7 (3.2), ACC: 7.3 (1.9))).  A comparison with traditional yet competitive ML methods shows that the transfer learning baselines XTab~\citep{zhu2023xtab} and CM2~\citep{Ye2024} consistently rank better in the AUROC scores. TransTab (AUROC: 4.8 (2.4), ACC: 4.6 (3.0)) outperforms all traditional ML methods based on ACC scores, but is outperformed by XGBoost (AUROC: 4.7 (1.6), ACC: 5.0 (2.9)) on AUROC scores.  In contrast, proposed CATTLE with self-supervised pretraining obtains the best average rank under both AUROC (2.9 (2.4)) and ACC (2.4 (1.5)) scores among all baseline methods.}

\subsection {Win matrix with statistical tests}

\begin{figure*}[t]
\centerline{\includegraphics[width=0.8\textwidth]{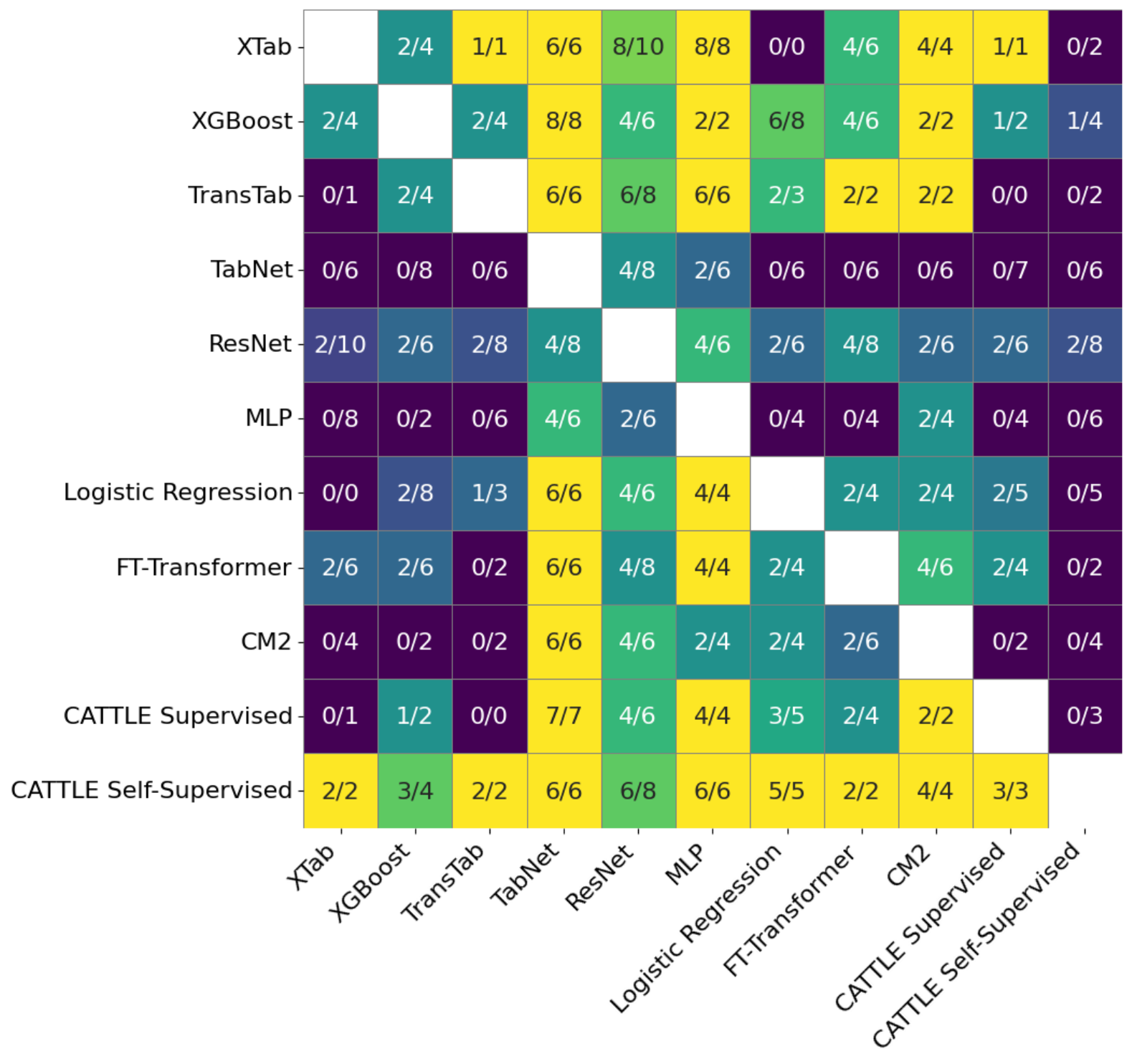}}
\vspace{-10pt}
\caption{Win matrix based on Wilcoxon signed-rank statistical tests. The cell value X/Y presents X as the number of times the row method statistically outperforms the column method. Y is the number of source–target data set pairs with statistically significant differences between a pair of methods.}
\label{fig-win-matrix}
\end{figure*}

\mds{A rank-based performance comparison of models in Section~\ref{ranking} presents an overall ordering of the methods without considering the magnitude or statistical significance of the improvement. We apply the Wilcoxon signed-rank test to evaluate whether the differences in AUROC scores between each pair of methods are statistically significant, thereby offering more detailed insights into performance gains.} 
The win matrix in Figure~\ref{fig-win-matrix} shows the statistical comparison between two methods in a pair. The method with the statistically higher AUROC score wins in the pair. Each cell in the win matrix shows the number of source-target data sets in which one method wins over the other. The win matrix shows that CATTLE with self-supervised pre-training wins against all the baseline methods on data sets that achieve statistical significance. For example, CATTLE with self-supervised pretraining and XGBoost produces statistically different AUROCs on four data sets, with CATTLE winning three out of four times.  In contrast, CATTLE with supervised pretraining does not show statistically significant differences from TransTab (0/0). However, CATTLE with supervised pretraining wins all significant cases against MLP (2/2), TabNet (7/7), and CM2 (2/2). XTab outperforms CATTLE supervised in a source-target data set pair (1/1).

\subsection{\mds{Comparison with state-of-the-art tabular models}}

\mds{We additionally benchmark CATTLE against two recently proposed state-of-the-art models for learning tabular data: TARTE~\citep{kim2025_tarte} and TabPFNv2~\citep{hollmann2025_tabpfnv2}. The learning strategy used by these two models is fundamentally different from that of CATTLE, since they rely extensively on large-scale data resources. TabPFNv2 is pre-trained on 130 million synthetic data sets, whereas TARTE uses a large volume of Wikipedia text data. Table~\ref{tab:sota-comparison} in ~\ref{app:sota_methods} compares CATTLE with TabPFNv2 and TARTE. In several data set pairs, self-supervised CATTLE outperforms TabPFNv2, although it falls short in other instances. In most cases, self-supervised CATTLE outperforms TARTE. Notably, the advantage of CATTLE lies in its generalized context learning approach, which requires only a single data source to remain competitive with large-scale pretrained tabular models, such as TabPFNv2 and TARTE.}

\subsection{Ablation study}

Several factors can impact CATTLE performance, \mds{which are investigated in four ablation studies.}

\begin{table*}[t]
\centering
\caption{Effects of transferring attention layers from the source pretrained transformer (gFTT) to the downstream transformer. Attention layers are numbered from bottom (0) to top (L), as \{0, 1, ..., L-1, L\} for the projection weight (W) matrices.}
\label{tab:gftt-transfer-strategies}
\vspace{5 pt}
\scalebox{0.55}{
\begin{tabular}{lccc|ccc}
\toprule

& \multicolumn{3}{c}{\textbf{Bottom Layers$\rightarrow$ Bottom Layers}} 
& \multicolumn{3}{c}{\textbf{Top Layers $\rightarrow$ Bottom Layers}} \\

\cmidrule(lr){2-4} \cmidrule(lr){5-7}

& \textbf{\begin{tabular}[c]{@{}c@{}}gFTT$_s$($W^0$)\\ $\rightarrow$ gFTT$_t$($W^0$)\end{tabular}} 
& \textbf{\begin{tabular}[c]{@{}c@{}}gFTT$_s$($W^{0,1}$) \\ $\rightarrow$ gFTT$_t$($W^{0,1}$)\end{tabular}} 
& \textbf{\begin{tabular}[c]{@{}c@{}}gFTT$_s$($W^0$) \\ $\rightarrow$ gFTT$_t$($W^{0,1}$)\end{tabular}} 

& \textbf{\begin{tabular}[c]{@{}c@{}}gFTT$_s$($W^L$)\\ $\rightarrow$ gFTT$_t$($W^0$)\end{tabular}} 
& \textbf{\begin{tabular}[c]{@{}c@{}}gFTT$_s$($W^L, W^{L-1}$) \\ $\rightarrow$ gFTT$_t$($W^{0,1}$)\end{tabular}} 
& \textbf{\begin{tabular}[c]{@{}c@{}}gFTT$_s$($W^L$) \\ $\rightarrow$ gFTT$_t$($W^{0,1}$) \\ {[}Proposed{]}\end{tabular}} \\

\midrule

\begin{tabular}[c]{@{}l@{}}Cardiovascular \\ $\rightarrow$ diabetes\end{tabular}
& 0.818 (0.06) & 0.816 (0.06) & 0.813 (0.06)
& 0.810 (0.05) & 0.818 (0.05) & \textbf{0.819 (0.05)} \\

\begin{tabular}[c]{@{}l@{}}credit-g \\ $\rightarrow$ diabetes\end{tabular}
& \textbf{0.821 (0.06)} & 0.819 (0.06) & 0.816 (0.06)
& 0.814 (0.06) & 0.814 (0.05) & 0.815 (0.06) \\

\begin{tabular}[c]{@{}l@{}}mfeat-fourier \\ $\rightarrow$ vehicle\end{tabular}
& \textbf{0.928 (0.01)} & 0.927 (0.01) & 0.926 (0.01)
& 0.928 (0.02) & 0.927 (0.01) & 0.927 (0.01) \\

\begin{tabular}[c]{@{}l@{}}optdigits \\ $\rightarrow$ vehicle\end{tabular}
& 0.933 (0.01) & 0.930 (0.01) & 0.925 (0.01)
& 0.931 (0.02) & 0.925 (0.01) & \textbf{0.942 (0.01)} \\

\begin{tabular}[c]{@{}l@{}}churn \\ $\rightarrow$ cmc\end{tabular}
& 0.745 (0.05) & \textbf{0.751 (0.05)} & 0.747 (0.05)
& 0.746 (0.03) & 0.747 (0.05) & 0.749 (0.05) \\

\begin{tabular}[c]{@{}l@{}}sick \\ $\rightarrow$ cmc\end{tabular}
& 0.745 (0.05) & 0.746 (0.04) & 0.747 (0.05)
& 0.748 (0.05) & 0.748 (0.05) & \textbf{0.748 (0.05)} \\

\begin{tabular}[c]{@{}l@{}}car-evaluation \\ $\rightarrow$ pc1\end{tabular}
& 0.836 (0.04) & 0.832 (0.04) & 0.838 (0.03)
& \textbf{0.844 (0.03)} & 0.843 (0.03) & 0.844 (0.05) \\

\begin{tabular}[c]{@{}l@{}}steel-plates-fault \\ $\rightarrow$ pc1\end{tabular}
& 0.833 (0.04) & 0.832 (0.04) & \textbf{0.845 (0.04)}
& 0.840 (0.03) & 0.834 (0.04) & 0.834 (0.04) \\

\begin{tabular}[c]{@{}l@{}}steel-plates-fault \\ $\rightarrow$ cylinder-bands\end{tabular}
& 0.844 (0.04) & 0.839 (0.04) & 0.841 (0.04)
& \textbf{0.853 (0.03)} & 0.847 (0.03) & 0.847 (0.03) \\

\begin{tabular}[c]{@{}l@{}}seismic-bumps \\ $\rightarrow$ cylinder-bands\end{tabular}
& 0.845 (0.03) & 0.843 (0.03) & 0.845 (0.03)
& 0.847 (0.03) & 0.856 (0.03) & \textbf{0.856 (0.03)} \\

\bottomrule
\end{tabular}}
\end{table*}

\begin{figure}[t]
\centerline{\includegraphics[width=0.9\textwidth]{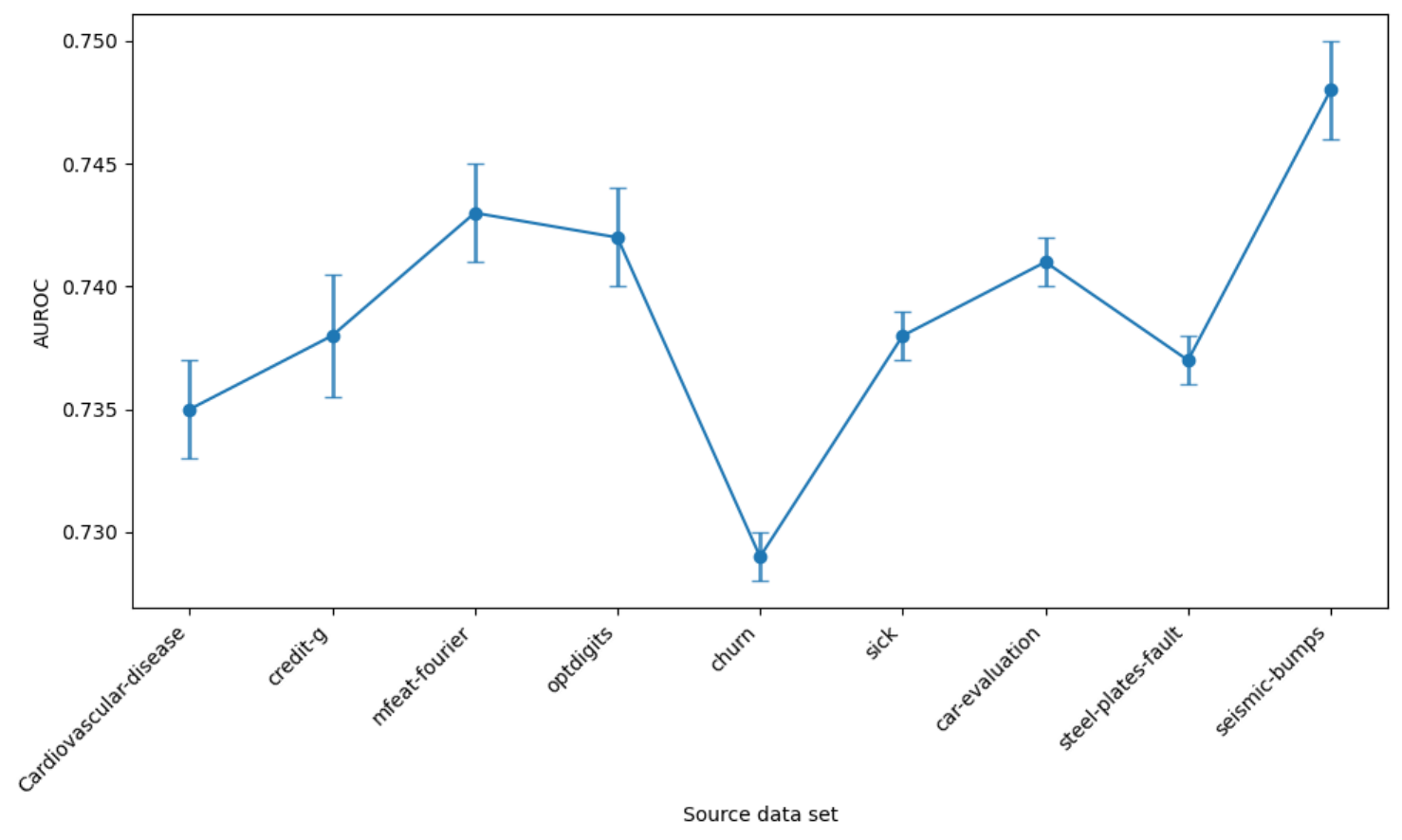}}
\caption{Effects of individual source data sets on the classification of the (downstream) cmc target data set. A source data set is used for self-supervised pretraining of the upstream model.
}
\label{fig-cattle-multi-source}
\end{figure}

First, \mds{the selection of weights from different attention layers of  the upstream source model may have impact on downstream performance. Table~\ref{tab:gftt-transfer-strategies} shows the effect of transferring projection weights from bottom and top attention layers of the upstream model. The performance differences between the choices of attention layers are marginal. The top-to-bottom configurations, which includes our proposed method, show achieves the best performance on six of the ten source-target pairs.} Second, the domain-specific context of the source data can affect the downstream classification of the target data set. Figure \ref{fig-cattle-multi-source} shows that the difference between the maximum and minimum AUROC scores due to the varying source data sets is 0.018, suggesting a negligible impact of the context specific to the source data. The improved performance of CATTLE, despite the insignificant effect of the domain-specific context, suggests the effectiveness of the generalized context.  
\mds{Third, Table~\ref{tab:ca-mechanism-ablation} shows the effect of randomly selecting, freezing or updating the selected weights. Instead of selecting weights from sequential layers in Figure~\ref{tab:gftt-transfer-strategies}, the random weights variant randomly selects the frozen key-value weights across all the layers of the pretrained model. The trained-weights variant transfers the proposed pretrained key-value projection weights but updates them during target fine-tuning. The direct fine-tuning variant directly fine-tunes the pretrained source gFTT model on the target data without requiring a separate downstream model.  Overall, CATTLE attains the highest performance on seven out of the ten source–target combinations, while traditional transfer learning via direct fine-tuning yields the best results on two of the pairs.} \mds {Fourth, Table~\ref{tab:embedding-ablation} presents the effect of BERT-based feature encoders compared to other conventional variants without context learning. In the anonymized variant, original feature names and categorical values are replaced with anonymous integer IDs, removing their semantics before learning the embedding using a linear projection layer. In the Word2Vec variant, feature names and categorical values are kept as text tokens, but their vector representations (embeddings) are randomly initialized and updated during training without using representations from a pretrained BERT model. The superiority of the BERT-based encoder suggests that preserving semantic and context information of feature names and categorical values in data representations consistently improves model performance. Therefore, BERT-based encoders have completely replaced conventional encoding methods in the current tabular data literature~\citep{wang2022transtab, Ye2024}.}

\begin{table*}[t]
\centering
\caption{\mds{Ablation study on categorical feature encoding strategies. The anonymized variant removes semantic information from feature names before learning their embeddings from scratch. The Word2Vec-style variant uses trainable token embeddings for feature names and categorical values. Embeddings for categorical feature names and values are extracted from pre-trained BERT~\citep{devlin2019_bert}.}}
\vspace{5 pt}
\label{tab:embedding-ablation}
\scalebox{0.80}{
\begin{tabular}{lccc}
\toprule
\textbf{Source $\rightarrow$ Target} & \textbf{Anonymized} & \textbf{Word2Vec} & \textbf{Pre-trained BERT} \\
\midrule

Cardiovascular-Disease-dataset $\rightarrow$ diabetes & 0.714 (0.06) & 0.707 (0.05) & \textbf{0.799 (0.04)} \\
credit-g $\rightarrow$ diabetes                       & 0.685 (0.04) & 0.694 (0.05) & \textbf{0.818 (0.04)} \\
mfeat-fourier $\rightarrow$ vehicle                   & 0.435 (0.06) & 0.441 (0.03) & \textbf{0.928 (0.01)} \\
optdigits $\rightarrow$ vehicle                       & 0.418 (0.05) & 0.432 (0.03) & \textbf{0.935 (0.01)} \\
churn $\rightarrow$ cmc                               & 0.582 (0.05) & 0.610 (0.04) & \textbf{0.726 (0.01)} \\
sick $\rightarrow$ cmc                                & 0.569 (0.06) & 0.601 (0.04) & \textbf{0.733 (0.02)} \\
car-evaluation $\rightarrow$ pc1                      & 0.689 (0.13) & 0.724 (0.06) & \textbf{0.825 (0.04)} \\
steel-plates-fault $\rightarrow$ pc1                  & 0.707 (0.07) & 0.684 (0.16) & \textbf{0.812 (0.04)} \\
steel-plates-fault $\rightarrow$ cylinder-bands       & 0.565 (0.08) & 0.498 (0.13) & \textbf{0.827 (0.03)} \\
seismic-bumps $\rightarrow$ cylinder-bands            & 0.572 (0.09) & 0.548 (0.10) & \textbf{0.825 (0.04)} \\

\bottomrule
\end{tabular}}
\end{table*}

\begin{table*}[t]
\centering
\caption{\mds{Ablation study on cross-attention weight transfer mechanisms. The $random$-$weights$ variant uses randomly selected weights from the pretrained model to remain frozen in the downstream. The $trained$-$weights$ variant updates the proposed frozen pre-trained $key$-$value$ projection weights in the downstream. The $direct$ $finetune$ variant directly finetunes the upstream pre-trained model on the target data without proposed weight transfer. Proposed CATTLE transfers selective pretrained key-value projection weights and keeps them frozen while finetuning the downstream model using target data.}}
\vspace{5 pt}
\label{tab:ca-mechanism-ablation}
\scalebox{0.7}{
\begin{tabular}{lcccc}
\toprule
\textbf{Source $\rightarrow$ Target} & Random & Trained & Direct & Proposed \\
&weights& weights& finetune & CATTLE\\
\midrule

Cardiovascular-Disease-dataset $\rightarrow$ diabetes & 0.812 (0.06) & 0.810 (0.06) & 0.799 (0.04) & \textbf{0.819 (0.05)} \\

credit-g $\rightarrow$ diabetes & 0.811 (0.06) & 0.815 (0.06) & \textbf{0.818 (0.04)} & 0.815 (0.06) \\

mfeat-fourier $\rightarrow$ vehicle & 0.901 (0.01) & 0.912 (0.01) & \textbf{0.928 (0.01)} & 0.927 (0.01) \\

optdigits $\rightarrow$ vehicle                       & 0.932 (0.01) & 0.927 (0.01) & 0.935 (0.01) & \textbf{0.942 (0.01)} \\

churn $\rightarrow$ cmc                               & 0.745 (0.04) & 0.747 (0.05) & 0.726 (0.01) & \textbf{0.749 (0.05)} \\

sick $\rightarrow$ cmc                                & 0.728 (0.05) & 0.735 (0.05) & 0.733 (0.02) & \textbf{0.748 (0.05)} \\

car-evaluation $\rightarrow$ pc1                      & 0.845 (0.04) & \textbf{0.846 (0.03)} & 0.825 (0.04) & 0.844 (0.05) \\

steel-plates-fault $\rightarrow$ pc1                  & 0.818 (0.04) & 0.832 (0.03) & 0.812 (0.04) & \textbf{0.834 (0.04)} \\

steel-plates-fault $\rightarrow$ cylinder-bands       & 0.838 (0.04) & 0.840 (0.03) & 0.827 (0.03) & \textbf{0.847 (0.03)} \\

seismic-bumps $\rightarrow$ cylinder-bands            & 0.847 (0.04) & 0.838 (0.03) & 0.825 (0.04) & \textbf{0.856 (0.03)} \\
\bottomrule
\end{tabular}}
\end{table*}

\section {Discussion} 
\label{sec: Discussion}

This article presents one of the first transfer learning methods for tabular data sets without shared features. Our research findings can be summarized as follows. First, cross-attention via transformer weights rather than $key$, $value$, and $query$ representations yields state-of-the-art performance in cross-domain transfer learning. Second, cross-attention at the transformer weight level yields generalized context independent of the upstream source context and representation. Third,  a single data source can provide a \emph{data-agnostic} generalized context for cross-domain learning without requiring large volumes of disjoint data sets from disparate domains. Fourth, self-supervised pretraining using unlabeled source data is better than its supervised counterpart for \emph{data-agnostic} transfer learning.  The key findings of this paper require further elucidation in the context of existing work.

\subsection {Traditional versus proposed transfer learning}

Traditional transfer learning fine-tunes the same pre-trained model under the assumption that source data knowledge and representations are informative and transferable to learning downstream target data. However, disjoint tabular data sets from disparate domains are unlikely to share a general feature representation for knowledge transfer. A model pre-trained on one domain may lack the appropriate context to target data from a different domain, leading to cross-domain knowledge conflicts. The proposed \emph{data-agnostic} learning minimizes cross-domain knowledge conflicts in two ways. First, the attention layer weights of the pretrained model are transferred to a new target model rather than fine-tuning the pretrained model downstream. Second, cross-domain attention is achieved using transformer weights instead of $key$ and $value$ representations. 

It can be argued that the core gFTT model for representation learning plays a major role in CATTLE performance. However, other baselines, including TransTab\citep{wang2022transtab} and CM2~\citep{Ye2024}, use gFTT or a similar feature-tokenized transformer. Furthermore, selecting attention layers from pretrained gFTT models in our approach may not align with other transfer learning frameworks. In addition to the ablation study, we demonstrate the efficacy of five gFTT attention layers used in this study. Table \ref{tab:gftt_active_layers} in \ref{app:gftt_active_layers} shows that the topmost layers of a pre-trained model are most effective for downstream classification performance. In line with this observation, the weights of the topmost pretrained attention layers are selected for downstream learning of target data in this paper.

\subsection {Single versus multi-source data for pretraining}
The choice of a single-source data set over large volumes of multi-source data may contradict the general practice of developing pretrained models for transfer learning. Large language and vision models are developed using data from various application domains. Thanks to shared image patterns and text semantics across domains, which facilitate seamless incremental learning of general knowledge from large volumes of data. Similar shared features are not plausible in medical records and banking transactions when represented in tabular form. Without proper inductive bias, training a model using disjoint data from disparate domains can compromise the integrity of knowledge, override knowledge due to domain conflicts, and lead to memorizing new domains rather than generalizing knowledge. A similar perspective can explain why the pretrained CM2 model, despite being trained on 2000 tabular data sets, underperforms baselines using either a single (TransTab) or much fewer than 2000 data sets (XTab). Therefore, data-agnostic transfer learning, preferably using a generalized context similar to ours, is imperative and effective for tabular data.

\subsection {Transfer learning for limited sample data sets}
In the context of tabular data with limited samples, traditional machine learning is recommended, as transfer learning is not trivial given the lack of effectively pretrained models. For example, the Cylinder Bands (CB) data set has the smallest sample size (540) and the highest number of features (39), with an even mix of numerical and categorical variables. Deep representation learning (ResNet, FT-Transformer) and even recent transfer learning methods (XTab, CM2, TransTab) fall short of the performance of the XGBoost classifier on the CB data set. However, CATTLE with self-supervised pre-training outperforms XGBoost, suggesting an effective solution to learning challenging tabular data with limited samples.

\subsection {Limitations}
The proposed method is not without limitations, despite demonstrating state-of-the-art performance. The selection of attention layers for weight transfer is based on prior knowledge about knowledge distribution in deep layers, which would require a theoretical underpinning. More research is needed to optimize this framework by adopting a more systematic approach to selecting the attention layers. It is challenging to clearly explain the effectiveness of \emph{data-agnostic} transfer learning without understanding the distribution of knowledge in multi-head attention layers. 

\section{Conclusions} \label{sec: Conclusion}

This article presents a novel cross-domain attention method that accounts for the inductive bias required for the transfer learning of heterogeneous tabular data. Attention at the transformer weight level, rather than in $key$ and $value$ representations, yields \emph{data-agnostic} generalized context necessary for cross-domain transfer learning. A single source data set can learn a generalized context for a downstream task in a different domain. The proposed generalized context in transfer learning outperforms state-of-the-art methods for classification and transfer learning on tabular data, as well as models pretrained on large tabular datasets.

\section{Acknowledgments}

Research reported in this publication was supported by the US National Science Foundation (NSF) award \# 2431058. The research reported in this publication received support partly by the Air Force Office of Scientific Research under Grant Number W911NF-23-1-0170. The content is solely the responsibility of the authors and should not be interpreted as representing the official policies, either expressed or implied, of the Army Research Office or the U.S. Government.

\bibliographystyle{elsarticle-num}
\bibliography{CATTLE}

\newpage

\appendix
\section{Appendix}

\subsection{Effects of the number of active attention layers}
\label{app:gftt_active_layers}
\begin{table*}[ht]
\centering
\caption{Effects of the number of active attention layers of the pretrained gFTT model (supervised) on downstream target data classification (AUROC scores).}
\label{tab:gftt_active_layers}
\scalebox{0.85}{
\begin{tabular}{cccccccc}
\toprule
Data set pairs       & Top 4 Layers & Top 3 Layers & Top 2 Layers & Top 1 Layer  \\
\midrule
CD $\rightarrow$ DB  & 0.818(0.051) & 0.814(0.055) & 0.810(0.060) & 0.821(0.057) \\
CE $\rightarrow$ PC1 & 0.787(0.071) & 0.810(0.051) & 0.833(0.047) & 0.831(0.035) \\
CH $\rightarrow$ CM  & 0.735(0.046) & 0.732(0.039) & 0.741(0.043) & 0.743(0.039) \\
CG $\rightarrow$ DB  & 0.803(0.063) & 0.808(0.059) & 0.806(0.063) & 0.804(0.061) \\
MF $\rightarrow$ VH  & 0.932(0.014) & 0.938(0.014) & 0.929(0.015) & 0.927(0.012) \\
DG $\rightarrow$ VH  & 0.928(0.010) & 0.929(0.010) & 0.926(0.010) & 0.923(0.012) \\
SK $\rightarrow$ CM  & 0.753(0.044) & 0.755(0.044) & 0.750(0.037) & 0.745(0.047) \\
SP $\rightarrow$ CB  & 0.842(0.042) & 0.846(0.029) & 0.841(0.040) & 0.838(0.032) \\
SP $\rightarrow$ PC1 & 0.814(0.041) & 0.805(0.044) & 0.820(0.044) & 0.846(0.032) \\
\bottomrule
\end{tabular}}
\end{table*}

\subsection{Performance against recent state-of-the-art models}
\label{app:sota_methods}
\begin{table*}[ht]
\centering
\caption{Comparison of CATTLE (supervised and self-supervised) with TARTE and TabPFNv2 across multiple source–target tasks.}
\label{tab:sota-comparison}
\scalebox{0.75}{
\begin{tabular}{cccccc}
\toprule
Target & TARTE & TabPFNv2 & Source & \begin{tabular}[c]{@{}c@{}}CATTLE\\ Supervised\end{tabular} & \begin{tabular}[c]{@{}c@{}}CATTLE\\ Self-Supervised\end{tabular} \\
\midrule
\multirow{2}{*}{DB} 
& \multirow{2}{*}{0.815 (0.04)} 
& \multirow{2}{*}{0.816 (0.03)} 
& CD & 0.803 (0.05) & 0.807 (0.05) \\
& & & CG & 0.808 (0.04) & \textbf{0.819 (0.04)} \\
\multirow{2}{*}{VH} 
& \multirow{2}{*}{0.941 (0.01)} 
& \multirow{2}{*}{\textbf{0.966 (0.01)}} 
& MF & 0.940 (0.01) & 0.936 (0.00) \\
& & & DG & 0.932 (0.01) & 0.942 (0.01) \\
\multirow{2}{*}{CM} 
& \multirow{2}{*}{0.737 (0.02)} 
& \multirow{2}{*}{\textbf{0.756 (0.02)}} 
& CH & 0.721 (0.03) & 0.729 (0.02) \\
& & & SK & 0.726 (0.03) & 0.738 (0.02) \\
\multirow{2}{*}{PC1} 
& \multirow{2}{*}{0.752 (0.06)} 
& \multirow{2}{*}{\textbf{0.902 (0.02)}} 
& CE & 0.842 (0.05) & 0.840 (0.08) \\
& & & SP & 0.845 (0.04) & 0.813 (0.05) \\
\multirow{2}{*}{CB} 
& \multirow{2}{*}{0.819 (0.04)} 
& \multirow{2}{*}{0.854 (0.02)} 
& SP & 0.829 (0.04) & \textbf{0.862 (0.04)} \\
& & & SB & 0.832 (0.03) & 0.850 (0.04) \\
\bottomrule
\end{tabular}}
\end{table*}

\end{document}